\documentclass[11pt]{article}

\usepackage[margin=1in]{geometry}

\usepackage{amsmath,amssymb}
\usepackage{tikz-cd}
\usetikzlibrary{positioning,arrows.meta}
\usepackage{pgfplots}
\pgfplotsset{compat=1.18}
\usepackage{multicol}
\usepackage{booktabs}
\usepackage{array}
\usepackage{caption}
\usepackage{graphicx}
\usepackage{xurl}
\usepackage[numbers,sort&compress]{natbib}
\usepackage[hidelinks]{hyperref}

\title{Evaluating Cloud and Self-Hosted OCR Systems for Arabic, English, and French Documents: Bounding Box Detection, Text Recognition, and Mathematical Expression Recognition}

\author{%
Jumana Alsubhi, Ahmed Alhusayni, Abdulrahman Gharawi, Israa Hamdine,\\
Khadija Alansari, Alshaymaa Allahim, Lamees Alhumaid, and Ahmad Shabana \\[0.8em]
\normalsize\textit{Naseej Innovation Lab, Naseej Technical Company, Riyadh, Saudi Arabia}\\[0.3em]
\small\texttt{\{Jumana.Alsubhi, Ahmad.Alhusainy, A.Gharawi, Israa.Hamdin,}\\
\small\texttt{Khadija.Alansari, A.Allahim, L.Alhumaid, ashabana \}@naseej.com}
}
\date{}

\begin{document}

\maketitle

\begin{abstract}
Choosing an OCR system for a multilingual archive requires balancing accuracy
against deployment constraints. This study compares cloud and self-hosted
systems on three tasks: bounding box detection on 48 Arabic, English, and French
pages, text recognition on 315 pages across nine document types, and
mathematical expression recognition on 83 English and Arabic samples. Detection
was assessed by page-level visual grades, text by character error rate (CER)
against a Gemini-generated reference transcription, and mathematics by scores
against known answers. Main estimates are accompanied by 95\% intervals and
paired comparisons. Google Document AI and Surya produced Good line boxes on
47 of 48 (98\%) and 46 of 48 (96\%) pages. Datalab, Azure Document Intelligence,
and Document AI had mean CERs of 0.145--0.152, with no difference detected in
unadjusted paired comparisons. Surya had the lowest mean CER among self-hosted systems
(0.199). No mathematical test had a statistically clear leader. Strict checks
showed that near-perfect page scores could conceal a replaced Arabic variable or
Eastern Arabic digits rewritten in Western form. Reliable evaluation therefore
requires uncertainty estimates, structural checks, and notation-specific tests
in addition to aggregate recognition scores.
\end{abstract}

\vspace{\baselineskip}

\begin{multicols}{2}

\section{Introduction}

Optical character recognition (OCR) is a core component of document
digitisation. It converts document images into machine-readable text for search,
indexing, and downstream information extraction
\cite{li2023trocr}.
Reliable document processing requires more than character transcription. A
system must locate text, preserve reading order and layout, and represent tables
and mathematical expressions in usable forms
\cite{ouyang2025omnidocbench,blecher2023nougat}. Errors in any of these stages can reduce
the reliability of the complete workflow.

Multilingual collections make these tasks more difficult. Arabic, English, and
French may occur in the same archive or page. Arabic adds challenges related to
right-to-left writing, connected letter forms, diacritics, mixed scripts,
Eastern Arabic digits, handwriting, calligraphy, and manuscripts
\cite{bhatia2024qalam,heakl2025kitab,ismail2023hicma}. Real documents may also be
faded, stained, skewed, photographed, or arranged in tables and multi-column
layouts. Performance on clean printed text therefore does not represent all
practical conditions \cite{heakl2025kitab}.

Contemporary OCR deployments include managed cloud services, general
vision-language models, specialised recognition models, and self-hosted
engines. Cloud and self-hosted systems differ in recognition performance and in
deployment constraints, including speed, hardware, cost, licensing, and data
residency. System selection must therefore consider both.

A single text-recognition score does not capture the full document-processing
task \cite{liu2023ocrbench}. Bounding box detection measures whether text is
located correctly. Text recognition measures transcription quality across
languages and document types. Mathematical recognition measures whether
equations retain their symbols, structure, variables, and notation
\cite{deng2017image,blecher2023nougat}. Aggregate scores may also hide serious
errors, including incomplete tables, incorrect reading order, and plausible but
incorrect substitutions in Arabic text and notation
\cite{heakl2025kitab,ismail2023hicma,ouyang2025omnidocbench}.

This study evaluates cloud and self-hosted OCR systems through three
complementary components. The bounding box detection benchmark compares five
systems on 48 real pages. The text recognition benchmark compares ten systems
on 315 Arabic, English, and French pages. The mathematical recognition benchmark
compares thirteen systems on 83 English and Arabic samples. The evaluations
combine quantitative scoring with qualitative error analysis and operational
measures.

The main contributions of this study are: (i) a structured benchmark covering
bounding box detection, multilingual text recognition, and mathematical
recognition; (ii) analysis across Arabic, English, French, and multiple document
types; (iii) evaluation of Arabic-specific challenges in text and mathematical
notation; (iv) comparison of cloud and self-hosted systems under practical
deployment constraints; and (v) statistical and diagnostic analysis using
confidence intervals, paired tests, robustness checks, and strict checks of
Arabic symbols and digits to reveal errors hidden by aggregate scores.

\section{Related Work}

\subsection{Document OCR and Understanding}

Modern document-processing research has moved from isolated character
recognition toward systems that combine text, layout, and visual information.
Tesseract represents the established pipeline approach, combining page analysis,
line finding, character classification, and adaptive recognition
\cite{smith2007tesseract}.

Recent models extend recognition through large-scale pretraining. TrOCR
formulates text recognition as an image-to-sequence task using pretrained
transformers \cite{li2023trocr}. Nougat extends image-to-sequence recognition to
complete academic pages containing prose and mathematical expressions
\cite{blecher2023nougat}. Together, these approaches motivate evaluating both
transcription and structured document output.

Large datasets have supported progress in layout and table analysis. DocLayNet
provides human-annotated page regions across diverse document types
\cite{pfitzmann2022doclaynet}, while PubTables-1M adds detailed table structure
and functional labels
\cite{smock2022pubtables}. These resources supply the structural labels on which
layout and table evaluation depend.

Recent benchmarks evaluate broader OCR capabilities. OCRBench covers text
recognition, document question answering, key information extraction, and
handwritten mathematical expression recognition \cite{liu2023ocrbench}.
OmniDocBench evaluates end-to-end
PDF parsing and individual components such as text, formulas, tables, and reading
order across varied document sources \cite{ouyang2025omnidocbench}. Their
task-level design, rather than a single overall score, motivates the evaluation
used in this paper.

\subsection{Arabic OCR}

Arabic OCR research has increasingly addressed the limitations of benchmarks
focused on clean printed text. Recent work includes Qalam for printed and
handwritten recognition \cite{bhatia2024qalam}, HICMA for calligraphy and
manuscripts \cite{ismail2023hicma}, and KITAB-Bench for multi-domain OCR and
document understanding \cite{heakl2025kitab}.

Together, these datasets show that handwriting, calligraphy, and historical
material must be evaluated separately from print. This motivates the division of
the text set by document type. Arabic mathematical notation and cloud versus
self-hosted deployment remain comparatively underrepresented.

\subsection{Mathematical Recognition}

Mathematical recognition differs from ordinary text recognition because the
output must preserve both symbols and two-dimensional structure. Image-to-markup
models generate LaTeX directly from formula images
\cite{deng2017image}. Nougat extends this approach to complete academic pages
containing text and mathematical expressions \cite{blecher2023nougat}.
The CROHME competitions established shared data and separate evaluation of
symbols, formula structure, and matrices for handwritten mathematics
\cite{mouchere2016crohme}. CROHME concerns online handwriting recorded as
strokes; it provides evaluation context, not source data for the image-based
benchmark used here.

Evaluation is also difficult because different LaTeX sequences can render the
same expression. Character Detection Matching addresses this issue by comparing
rendered formulas at the image level rather than relying only on textual edit
distance \cite{wang2025cdm}. Existing work has advanced printed and handwritten
formula recognition. In the literature reviewed for this study, however, Arabic
variables, Eastern Arabic digits, and Arabic function notation appear less
represented in widely used mathematical OCR evaluations
\cite{liu2023ocrbench,mouchere2016crohme,ouyang2025omnidocbench}.

\subsection{Research Gap}

Prior work provides strong benchmarks for layout analysis, multilingual OCR,
Arabic documents, tables, and mathematical expressions. However, these areas usually emphasise different tasks or document collections. This study
complements that work through a practical comparison of bounding box detection,
multilingual text recognition, and mathematical recognition, with explicit
diagnostics for Arabic notation and attention to deployment constraints. The
contribution is this combination of tasks and diagnostics, rather than a claim
that no earlier benchmark covers related capabilities.

\section{Benchmark Construction}

\subsection{Benchmark Overview}

The benchmark contains three complementary components: bounding box detection,
multilingual text recognition, and mathematical recognition. Each component was
constructed around its own task requirements and evaluation unit. The detection
and text-recognition sets use full document pages, while the mathematical set
combines cropped formulas, generated questions, and complete pages.

The samples were purposefully selected to cover realistic variation rather than
to represent a single document source. Coverage includes clean and degraded
print, handwriting, manuscripts, calligraphy, tables, forms, receipts,
mixed-script pages, and mathematical notation. This design follows the broader
shift toward document-type and capability-level evaluation in OCR benchmarks
\cite{liu2023ocrbench,heakl2025kitab,ouyang2025omnidocbench}.

\subsection{Bounding Box Detection Set}

The bounding box detection set contains 48 pages in Arabic, English, and French,
and all five systems were evaluated on every page (Table~\ref{tab:bbox-set}).
The pages were selected to test line localisation across different scripts,
layouts, and image conditions. Arabic and French pages emphasise difficult
cases, while the larger English subset provides broader coverage of receipts,
forms, reports, and dense small print.

\begin{center}
\captionof{table}{Bounding box detection set.}
\label{tab:bbox-set}
\small
\begin{tabular}{@{}lr@{}}
\toprule
Language & Pages \\
\midrule
Arabic & 8 \\
French & 6 \\
English & 34 \\
\midrule
Total & 48 \\
\bottomrule
\end{tabular}
\end{center}

The Arabic pages include receipts, a newspaper, a handwritten invoice, a visa
form, and a university certificate. The French pages include forms,
prescriptions, receipts, and a statistical report. The English pages include
retail and tax receipts, business forms, reports, and fax sheets. Several pages
contain handwriting, mixed Arabic--English text, curved photographs, faded
thermal print, tables, charts, or multi-column layouts. These characteristics
extend the structural variation represented in document-layout benchmarks
\cite{pfitzmann2022doclaynet}.

\subsection{Text Recognition Set}

The text recognition set contains 315 pages divided into nine document types (Table~\ref{tab:text-set}).
Arabic accounts for 174 pages, French for 105, and English for 36. The
distribution gives particular attention to Arabic material and to document
types that remain difficult for OCR, including handwriting, manuscripts,
calligraphy, and tables.

\vspace{1em}

\begin{center}
\captionof{table}{Text recognition set by document type.}
\small
\label{tab:text-set}
\begin{tabular}{@{}lr@{}}
\toprule
Document type & Pages \\
\midrule
Arabic print & 119 \\
Arabic handwriting & 10 \\
Arabic manuscripts & 10 \\
Arabic calligraphy & 13 \\
Arabic tables & 22 \\
English print & 26 \\
English tables & 10 \\
French print & 86 \\
French tables & 19 \\
\midrule
Total & 315 \\
\bottomrule
\end{tabular}
\end{center}

The printed subsets contain books, reports, manuals, magazines, and official
documents from modern and historical sources. The table subsets contain
statistical, financial, and research tables with dense rows, small print, and
complex alignment. The Arabic categories reflect challenges identified in
recent Arabic OCR datasets, particularly mixed scripts, historical material,
handwriting, and calligraphy \cite{bhatia2024qalam,ismail2023hicma,heakl2025kitab}.

\subsection{Mathematical Recognition Set}

The mathematical recognition set contains 83 samples across five tests (Table~\ref{tab:math-set}). Every
sample has a known reference answer, allowing direct accuracy measurement. The
set combines controlled samples with real pages so that formula transcription
can be evaluated separately from full-page reading.

\begin{center}
\captionof{table}{Mathematical recognition set.}
\small
\label{tab:math-set}
\begin{tabular}{@{}p{0.58\columnwidth}r@{}}
\toprule
Test & Samples \\
\midrule
English formulas & 35 \\
Arabic pages with equations & 30 \\
Arabic notation & 10 \\
Real English pages & 7 \\
Real Arabic page & 1 \\
\midrule
Total & 83 \\
\bottomrule
\end{tabular}
\end{center}

The English formula subset contains 35 cropped expressions with verified LaTeX
labels from LaTeX\_OCR and im2latex-100k: 20 printed formulas (10 from each
source) and 15 handwritten formulas from LaTeX\_OCR, of which 8 use synthetic
handwriting and 7 were written by people \cite{deng2017image,latexocrdataset}.
The generated
Arabic subset contains 30 questions with equations written in standard Latin
notation. A further 10 generated expressions use Eastern Arabic digits, Arabic
variables, and Arabic function names. The eight real pages include printed and
handwritten mathematics, matrices, a skewed low-contrast scan, and an Arabic
page mixing Arabic labels with equations. This combination reflects both
image-to-LaTeX recognition and complete-page parsing settings
\cite{deng2017image,blecher2023nougat}.

\subsection{Construction Principles}

The three components follow four common principles. First, they include
multiple languages, scripts, and document types. Second, they combine routine
pages with deliberately difficult cases. Third, they retain task-specific
references: visual page review for detection, a reference transcription produced
by Gemini 3.1 Flash-Lite for text (chosen for its ability to reconstruct degraded
and historical text), and verified or constructed answers for mathematical
samples.
Fourth, they preserve operational diversity by evaluating both cloud and
self-hosted systems.

The benchmark is diagnostic rather than population-representative. Its purpose
is to expose differences between systems and identify failure patterns across
important document conditions.

\section{Methodology}

The evaluation framework was designed to compare heterogeneous OCR systems under
consistent task conditions. Within each benchmark component, systems received
the same source images and were evaluated using the same task-specific criteria.
Outputs were scored without manual correction. Only predefined normalisation,
such as removing markup before text comparison, was applied. This design follows
the controlled evaluation approach used in recent OCR benchmarks
\cite{liu2023ocrbench,heakl2025kitab,ouyang2025omnidocbench}. Figure~\ref{fig:workflow} summarizes the workflow, and Table~\ref{tab:components} lists the systems, samples, and scoring reference for each component.

\begin{center}
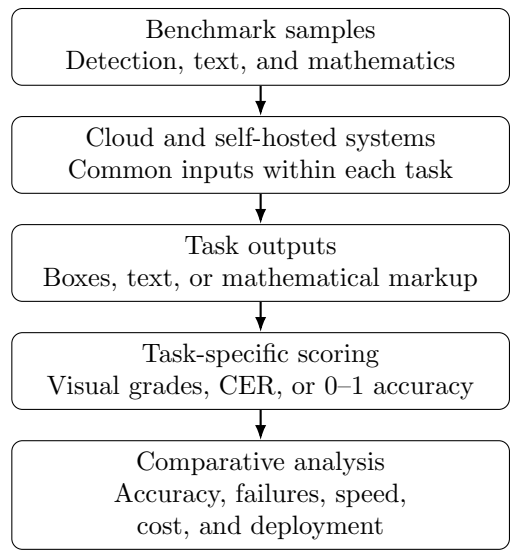

\begin{tikzpicture}[
    node distance=4mm,
    process/.style={draw, rounded corners, align=center,
        text width=0.78\columnwidth, inner sep=4pt, font=\small},
    flow/.style={-{Latex[length=2mm]}, thick}
]
\node[process] (samples) {Benchmark samples\\Detection, text, and mathematics};
\node[process, below=of samples] (systems) {Cloud and self-hosted systems\\Common inputs within each task};
\node[process, below=of systems] (outputs) {Task outputs\\Boxes, text, or mathematical markup};
\node[process, below=of outputs] (scores) {Task-specific scoring\\Visual grades, CER, or 0--1 accuracy};
\node[process, below=of scores] (analysis) {Comparative analysis\\Accuracy, failures, speed, cost, and deployment};
\draw[flow] (samples) -- (systems);
\draw[flow] (systems) -- (outputs);
\draw[flow] (outputs) -- (scores);
\draw[flow] (scores) -- (analysis);
\end{tikzpicture}
\captionof{figure}{Overall evaluation workflow.}
\label{fig:workflow}
\end{center}

\begin{center}
\captionof{table}{Summary of the evaluation components.}
\scriptsize
\label{tab:components}
\setlength{\tabcolsep}{3pt}
\begin{tabular}{@{}>{\raggedright\arraybackslash}p{0.23\columnwidth}cc>{\raggedright\arraybackslash}p{0.28\columnwidth}@{}}
\toprule
Component & Systems & Samples & Scoring reference \\
\midrule
Bounding box detection & 5 & 48 & Visual grades \\
Text recognition & 10 & 315 & Gemini reference; CER \\
Mathematical recognition & 13 & 83 & Known answers; score 0--1 \\
\bottomrule
\end{tabular}
\end{center}

\subsection{Evaluated Systems}

The bounding box evaluation compared five systems: Tesseract
\cite{smith2007tesseract}, Surya \cite{surya}, Qwen3-VL (8B) \cite{qwen3vl},
Google Document AI \cite{documentai}, and Gemini 3.1 Flash-Lite \cite{gemini}.
The text recognition
evaluation compared ten systems: Datalab, Azure Document Intelligence, Google
Document AI, Surya, olmOCR 2, Qwen3-VL, dots.ocr, PaddleOCR-VL, KATIB, and
Tesseract. The mathematical evaluation included these systems where applicable,
with Gemini 3.1 Flash-Lite, GOT-OCR 2.0 \cite{wei2024got}, and pix2tex, for a
total of thirteen.

Technical reports, official documentation, or model cards were used to identify
the evaluated systems and versions \cite{datalab,azuredi,olmocr2,dotsocr,
paddleocrvl,katib,pix2tex}.

The model set combines managed cloud services, specialised OCR models, general
vision-language models, and established OCR engines. Self-hosted systems were
run on one NVIDIA RTX A6000 (48~GB), while cloud systems were accessed through
managed APIs. Task-specific modes were used where available, including line
detection, layout extraction, formula detection, and LaTeX generation. Model
identifiers, versions, prompts, precision, and decoding settings are listed in
Appendix~A.

\subsection{Bounding Box Evaluation}

Each system was asked to locate text lines and return bounding coordinates.
Qwen3-VL returned $[x_1,y_1,x_2,y_2]$ and Gemini returned
$[y_{\min},x_{\min},y_{\max},x_{\max}]$, both on a 0--1000 scale. Coordinates
were converted to pixels of the original page. The convention was checked by
testing whether boxes covered page ink, remained within the page, and were not
predominantly vertical. Every page was then graded from a side-by-side panel
showing all systems on the same page:

\begin{itemize}
    \item \textit{Good}: boxes were tight, correctly placed, and covered almost
    all text.
    \item \textit{Usable}: most text was located, but some boxes were loose,
    merged, fragmented, or spurious.
    \item \textit{Poor}: substantial text was missed or boxes were severely
    misplaced.
\end{itemize}

The main measure was the proportion of pages graded \textit{Good}; the share
graded \textit{Good} or \textit{Usable} was also reported. As a complementary
automatic measure, ink F1 is the harmonic mean of ink recall (the share of page
ink covered by boxes) and box precision (the share of box area containing ink).

\subsection{Text Recognition Evaluation}

The text recognition set contained 315 pages. No human transcription existed for
most pages, so Gemini 3.1 Flash-Lite produced the reference transcriptions and
was excluded from the ten evaluated text-recognition systems. A general
vision-language model was chosen to provide a common reference for degraded,
handwritten, and historical material where human transcriptions were unavailable.
Its ability to reconstruct plausible text was not independently validated and
must not be treated as evidence of faithful transcription. Reconstruction can
introduce unsupported content or normalise spelling, digits, and punctuation,
potentially favouring systems with similar behaviour. The ranking was also
recomputed under four stricter normalisations.
These checks assessed sensitivity to text preparation; they did not validate
the reference against the source images.

Before comparison, reference and output text were normalised to Unicode NFC,
whitespace was collapsed, and HTML and Markdown markup was removed. These steps
evaluate recognised content rather than output formatting.

Text accuracy was measured using character error rate (CER):

\begin{equation}
\mathrm{CER}=\frac{S+D+I}{N},
\end{equation}

where $S$, $D$, and $I$ are character substitutions, deletions, and insertions,
and $N$ is the number of characters in the reference. Lower values indicate
closer agreement. CER is widely used for OCR evaluation, including multilingual
and Arabic benchmarks \cite{liu2023ocrbench,heakl2025kitab}.

Four aggregate measures were reported over the 315 scored pages: mean CER;
median CER, representing a typical page; capped mean CER, in which each page's
CER was limited to 1 so that a runaway output could not dominate the mean; and
the failed-page rate, defined as the share of outputs with at least one of the
six output problems specified in Appendix~\ref{app:normalisation}: empty output,
caption-like output, repetition loops, very short output, language switching,
or missing Arabic text. The results measure agreement with the
model-generated reference, not certified character-level accuracy.

\subsection{Mathematical Recognition Evaluation}

Mathematical outputs were scored against known reference answers on a scale from
0 to 1, where 1 indicates a complete match. Expressions were divided into
meaningful units, including commands, symbols, variables, operators, and
superscripts. Equivalent representations of the same symbol were normalised
before comparison. This token-based procedure gives partial credit while
preserving sensitivity to structural errors. It complements image-level formula
metrics designed to handle equivalent LaTeX renderings \cite{wang2025cdm}.

Scoring differed by test. English formulas were compared with reference LaTeX.
Arabic pages received the mean of a text-recognition score and an equation score.
Arabic-notation samples were scored for equation agreement. Three additional
strict checks recorded whether every symbol was kept, whether Arabic letters
used as symbols were kept, and whether Eastern Arabic numbers were retained as
written. Real pages were compared as complete outputs, including prose and
equations. Scores depended on agreement after the predefined token and markup
normalisation, rather than on a separately specified penalty for plain-text
output. A readable plain-text expression could still differ from its reference
in structural tokens. Removal of delimiters and layout commands did not
guarantee equivalence of every mathematically equivalent representation
\cite{deng2017image,blecher2023nougat}. Different tests used different metrics,
so scores and rankings should be compared within a test, not across rows.
Test-specific scores and additional Arabic-notation diagnostics are detailed in
Appendix~\ref{app:normalisation}.

All thirteen systems were scored on every mathematical test. An equation with
its two sides swapped was treated as a mismatch because the document was not
reproduced as written.

\subsection{Operational Measures}

The evaluation also recorded approximate processing time, GPU memory, usage
cost, licensing conditions, and processing location. Runtime was measured one
page or sample at a time in the test environment and should not be interpreted
as optimised batch throughput. GPU memory records the approximate peak observed
for self-hosted systems. Cloud costs were calculated from provider prices, while
self-hosted costs were treated as infrastructure usage rather than per-page
fees. Data residency records whether pages remained on local infrastructure or
were sent to an external service.

These measures allow accuracy to be interpreted alongside deployment
requirements.

\subsection{Statistical Analysis}

Confidence intervals are 95\% percentile bootstrap intervals over samples using
2,000 resamples and a fixed seed of 42. Differences between two systems were
assessed with a paired bootstrap on the samples processed by both systems; a
difference was called clear when its 95\% interval excluded zero. Ranking
stability under alternative text normalisations was measured with Kendall's
$\tau$. For bounding box grades, intervals on the share of \textit{Good} pages
are Wilson intervals.

These intervals and paired comparisons were exploratory and unadjusted for
multiple comparisons. An interval containing zero indicated that a difference
was not detected, not that two systems were statistically equivalent. Pages
were resampled individually; document-source clustering was not incorporated.
The pooled mean was a page-weighted summary of the curated collection, not a
language-balanced estimate or an estimate for all institutional collections.

\section{Results}

\subsection{Bounding Box Detection}

Document AI and Surya produced the strongest detection results: \textit{Good}
boxes on 47 (98\%) and 46 (96\%) of the 48 pages (Table~\ref{tab:detection-results}), with every remaining page
\textit{Usable}. Both were \textit{Good} on all 34 English pages. On the 14
Arabic and French pages, Document AI produced 13 \textit{Good} and one
\textit{Usable} result, and Surya produced 12 \textit{Good} and two
\textit{Usable} results.

\noindent\begin{minipage}{\linewidth}
\centering
\captionof{table}{Bounding box detection results by language. Intervals are 95\%
Wilson intervals for the share of \textit{Good} pages.}
\label{tab:detection-results}
\scriptsize
\setlength{\tabcolsep}{3pt}
\resizebox{0.90\linewidth}{!}{%
\begin{tabular}{@{}l>{\raggedright\arraybackslash}p{0.22\columnwidth}>{\raggedright\arraybackslash}p{0.22\columnwidth}cc@{}}
\toprule
System & English (34) & Arabic \& French (14) & Good & Good/usable \\
\midrule
Document AI & 34 G & 13 G, 1 U & 47/48, 98\% [89,100] & 48/48 \\
Surya & 34 G & 12 G, 2 U & 46/48, 96\% [86,99] & 48/48 \\
Gemini & 18 G, 15 U, 1 P & 10 G, 4 U & 28/48, 58\% [44,71] & 47/48 \\
Tesseract & 20 G, 14 U & 4 G, 4 U, 6 P & 24/48, 50\% [36,64] & 42/48 \\
Qwen3-VL & 6 G, 22 U, 6 P & 0 G, 2 U, 12 P & 6/48, 13\% [6,25] & 30/48 \\
\bottomrule
\end{tabular}
}
\end{minipage}

Gemini located text on 47 pages but produced line-level \textit{Good} boxes on
only 28 (58\%), because it often returned one box per paragraph or multi-line
cell. Its automatic ink F1 was nevertheless 0.795, nearly identical to Surya's
0.796. Tesseract was \textit{Good} on 24 pages and was weakest on handwriting,
low-resolution forms, and dense bilingual layouts. Qwen3-VL was \textit{Good}
on 6 pages and \textit{Poor} on 12 of the 14 Arabic and French pages. For the
four systems run through the common pipeline, median detection time was 0.16
s/page for Surya, 0.65 for Tesseract, 5.6 for Gemini, and 56.9 for Qwen3-VL.

Figure~\ref{fig:french-detection} illustrates these differences on the French
statistics report (FR\_006).

\end{multicols}
\begin{center}
\begin{minipage}{\textwidth}
\centering
\includegraphics[width=0.8\linewidth]{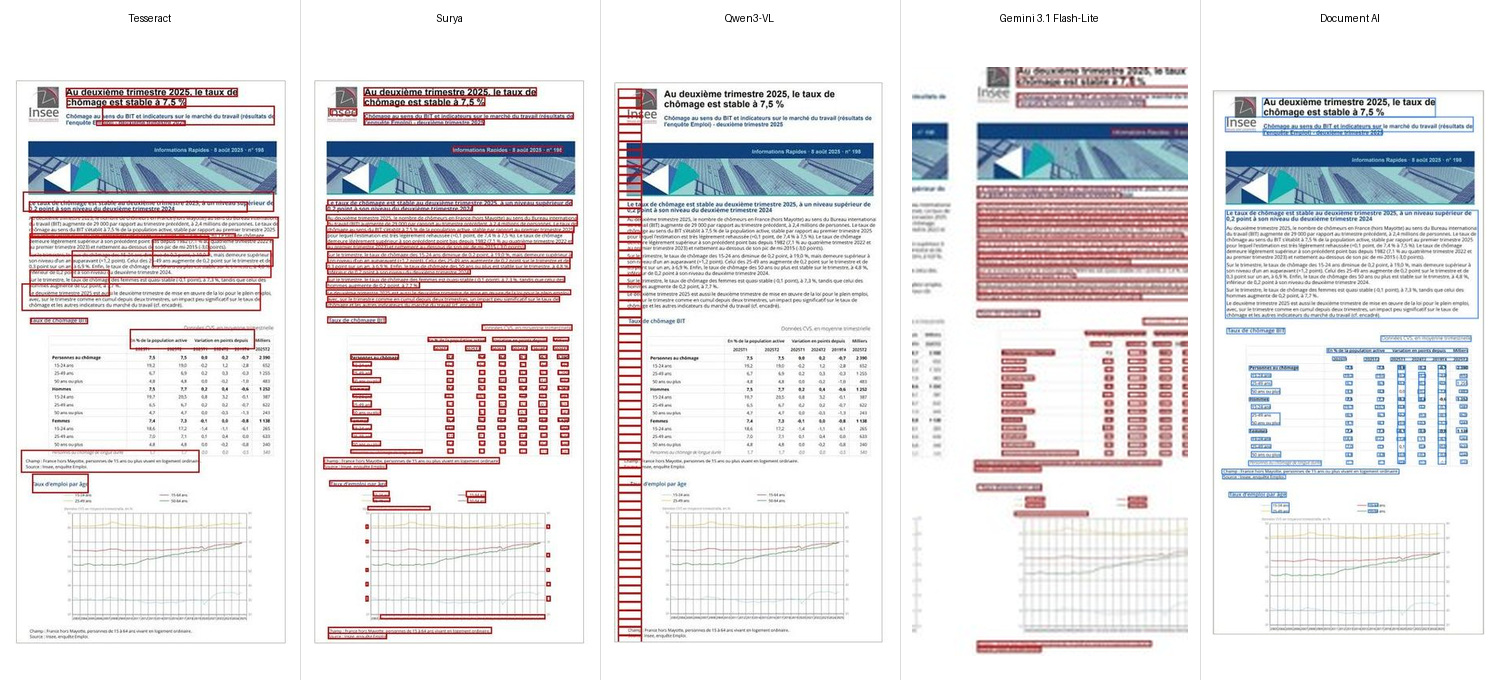}
\captionof{figure}{Bounding box detection on the French statistics report
(FR\_006). From left to right: Tesseract, Surya, Qwen3-VL, Gemini 3.1
Flash-Lite, and Document AI. }
\label{fig:french-detection}
\end{minipage}
\end{center}
\begin{multicols}{2}

\subsection{Text Recognition}

The three cloud services achieved the lowest mean CER, with no difference
detected between them in the unadjusted paired comparisons: Datalab
0.145 [0.120, 0.172], Azure DI 0.145 [0.122, 0.171], and Document
AI 0.152 [0.128, 0.176] (Table~\ref{tab:text-overall}). Surya had the lowest mean
among self-hosted systems at 0.199 [0.158, 0.251]. In paired comparisons, Surya
had higher mean CER than Datalab and Document AI, with unadjusted paired
intervals excluding zero; its 0.042 advantage over olmOCR 2 was not detected
as a difference (95\% interval $-0.002$ to 0.080; unadjusted $p=0.07$).

Figure~\ref{fig:text-cer} compares mean CER and its uncertainty with the median
and capped mean for each system.

\end{multicols}
\begin{center}
\begin{minipage}{\textwidth}
\centering
\includegraphics[width=0.85\linewidth]{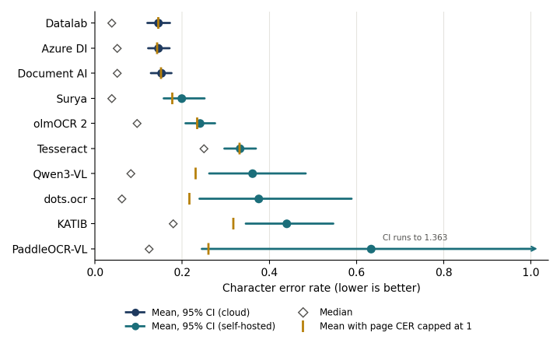}
\captionof{figure}{Text recognition CER against the model-generated reference
(lower is better). Circles show means with 95\% bootstrap confidence intervals,
diamonds show medians, and vertical ticks show means with page CER capped at 1.
The upper confidence limit for PaddleOCR-VL is 1.363 and extends beyond the
plotted range.}
\label{fig:text-cer}
\end{minipage}
\end{center}
\begin{table*}[t]
\centering
\small
\begin{tabular}{lrrrr}
\toprule
System & Mean CER & 95\% CI & Median & Capped \\
\midrule
Datalab & 0.145 & [0.120, 0.172] & 0.038 & 0.145 \\
Azure DI & 0.145 & [0.122, 0.171] & 0.051 & 0.143 \\
Document AI & 0.152 & [0.128, 0.176] & 0.051 & 0.152 \\
Surya & 0.199 & [0.158, 0.251] & 0.038 & 0.177 \\
olmOCR 2 & 0.241 & [0.208, 0.276] & 0.096 & 0.235 \\
Tesseract & 0.333 & [0.297, 0.369] & 0.250 & 0.332 \\
Qwen3-VL & 0.361 & [0.262, 0.483] & 0.082 & 0.231 \\
dots.ocr & 0.376 & [0.240, 0.589] & 0.061 & 0.217 \\
KATIB & 0.440 & [0.346, 0.547] & 0.179 & 0.317 \\
PaddleOCR-VL & 0.634 & [0.245, 1.363] & 0.124 & 0.260 \\
\bottomrule
\end{tabular}
\captionof{table}{Text recognition: mean CER with 95\% bootstrap interval, median, and mean with page CER capped at 1.}
\label{tab:text-overall}
\end{table*}
\begin{multicols}{2}

Mean CER was sensitive to runaway outputs. Capping every page at CER 1 reduced
PaddleOCR-VL from 0.634 to 0.260 and dots.ocr from 0.376 to 0.217. The ranking
was stable under letters-only comparison, removal of diacritics, equivalence of
Eastern and Western digits, and removal of punctuation (Kendall's
$\tau=0.96$--1.00 against the base ranking).

Failure-type analysis distinguished language switching from repetition and empty
outputs (Table~\ref{tab:failure-types}). PaddleOCR-VL had the highest failed-page
rate (9.5\%), followed by dots.ocr (7.0\%). Language switching affected 6.3\% of
PaddleOCR-VL pages, while repetition loops affected 5.1\% of KATIB pages and
4.1\% of dots.ocr pages. Categories overlapped, so their percentages were not
additive.

Table~\ref{tab:text-by-type} reports mean CER by document type. Datalab had the lowest CER on the four Arabic text types, with unadjusted paired
intervals excluding zero only on manuscripts (0.225 versus 0.339 for Document
AI) and calligraphy (0.399 versus 0.544). No difference was detected between the
leading cloud systems on print or handwriting. Surya was
the lowest-CER self-hosted system on all four. In exploratory comparisons,
Azure DI led Arabic tables
(0.173 versus 0.322 for Tesseract) and had the lowest CER on French tables
(0.049), although its difference from PaddleOCR-VL (0.081) was not detected.
Qwen3-VL had the lowest CER on English tables (0.171), with an unadjusted paired
interval excluding zero against dots.ocr on their common scored pages.

\end{multicols}
\begin{table*}[t]
\centering
\scriptsize
\resizebox{\textwidth}{!}{%
\begin{tabular}{lrrrrrrrrrr}
\toprule
 & Datalab & Azure DI & Document AI & Surya & olmOCR 2 & Tesseract & Qwen3-VL & dots.ocr & KATIB & PaddleOCR-VL \\
\midrule
AR-PRINT (119) & \textbf{0.129} & 0.142 & 0.139 & 0.203 & 0.225 & 0.368 & 0.363 & 0.602 & 0.373 & 0.326 \\
AR-HAND (10) & \textbf{0.170} & 0.171 & 0.177 & 0.251 & 0.373 & 0.759 & 0.504 & 0.483 & 0.307 & 0.444 \\
AR-MANU (10) & \textbf{0.225} & 0.342 & 0.339 & 0.381 & 0.738 & 0.874 & 1.217 & 0.831 & 0.658 & 0.775 \\
AR-CALLI (13) & \textbf{0.399} & 0.610 & 0.544 & 0.639 & 0.743 & 0.944 & 2.459 & 0.909 & 0.719 & 0.842 \\
AR-TABLE (22) & 0.327 & \textbf{0.173} & 0.410 & 0.374 & 0.585 & 0.322 & 0.576 & 0.434 & 0.631 & 0.488 \\
EN-PRINT (26) & 0.067 & 0.049 & 0.035 & 0.057 & \textbf{0.013} & 0.042 & 0.018 & 0.035 & 0.201 & 0.084 \\
EN-TABLE (10) & 0.347 & 0.402 & 0.316 & 0.356 & 0.621 & 0.395 & \textbf{0.171} & 0.224 & 0.568 & 0.395 \\
FR-PRINT (86) & 0.081 & 0.066 & \textbf{0.048} & 0.096 & 0.054 & 0.195 & 0.056 & 0.056 & 0.522 & 1.386 \\
FR-TABLE (19) & 0.098 & \textbf{0.049} & 0.108 & 0.118 & 0.223 & 0.190 & 0.086 & 0.211 & 0.283 & 0.081 \\
\bottomrule
\end{tabular}
}
\captionof{table}{Mean CER by document type against the model-generated reference (lower is better; pages in brackets). Bold marks the lowest value in each row.}
\label{tab:text-by-type}
\end{table*}
\begin{multicols}{2}

CER also understated structural incompleteness. olmOCR 2 returned less than half
of the reference text on 31\% of the 51 table pages, compared with 3\% of other
pages; Azure DI and Document AI did not do so on any table page. Values detached
from row labels were likewise not penalised when all characters remained.

\subsection{Mathematical Recognition}

No system had the highest score on every mathematical test (Table~\ref{tab:maths-grid}). Where paired
comparisons were possible, no difference was detected between the highest
scoring system and its runner-up (Table~\ref{tab:math-best}). This did not
establish equivalence; the single real Arabic page allowed only a descriptive
comparison.

\begin{center}
\begin{minipage}{\columnwidth}
\centering
\captionof{table}{Best cloud and self-hosted mathematical results. Intervals are
95\% bootstrap intervals. Available paired comparisons did not detect a
leader--runner-up difference; the single real Arabic page is descriptive.}
\label{tab:math-best}
\scriptsize
\setlength{\tabcolsep}{3pt}
\begin{tabular}{@{}>{\raggedright\arraybackslash}p{0.27\columnwidth}>{\raggedright\arraybackslash}p{0.29\columnwidth}>{\raggedright\arraybackslash}p{0.29\columnwidth}@{}}
\toprule
Test & Best cloud & Best self-hosted \\
\midrule
English formulas & Gemini 0.915 [0.875,0.949] & PaddleOCR-VL 0.849 [0.759,0.922] \\
Arabic pages & Datalab 0.998 [0.996,1.000] & Surya 0.997 [0.993,0.999] \\
Arabic notation & Azure DI 0.975 [0.948,1.000] & olmOCR 2 0.832 [0.702,0.925] \\
Real English pages & Gemini 0.998 [0.996,1.000] & olmOCR 2 1.000 [1.000,1.000] \\
Real Arabic page & Gemini 0.993 & olmOCR 2/Surya 0.993 \\
\bottomrule
\end{tabular}
\end{minipage}
\end{center}

On the seven real English pages, olmOCR 2 (1.000), dots.ocr and Surya (0.999),
Gemini (0.998), and Datalab (0.994) were within 0.006. On the real Arabic page,
Gemini, olmOCR 2, and Surya shared the highest score (0.993). Arabic notation had
the largest spread: Azure DI scored 0.975 and kept every symbol in 8 of 10
samples, but Document AI at 0.877 was not clearly behind. The best self-hosted
score was 0.832 from olmOCR 2, although it did not keep every symbol in any of
the 10 notation samples.

\end{multicols}
\begin{table*}[t]
\centering
\scriptsize
\resizebox{\textwidth}{!}{%
\begin{tabular}{lrrrrrrrrrrrrr}
\toprule
 & Gemini & Datalab & PaddleOCR-VL & dots.ocr & olmOCR 2 & Surya & Qwen3-VL & GOT-OCR 2.0 & pix2tex & Azure DI & KATIB & Document AI & Tesseract \\
\midrule
Formulas & \textbf{0.915} & 0.911 & 0.849 & 0.806 & 0.799 & 0.795 & 0.769 & 0.611 & 0.605 & 0.440 & 0.438 & 0.331 & 0.180 \\
Arabic pages & 0.996 & \textbf{0.998} & 0.252 & 0.995 & 0.990 & 0.997 & 0.823 & 0.088 & 0.020 & 0.489 & 0.737 & 0.654 & 0.365 \\
Arabic notation & 0.358 & 0.553 & 0.055 & 0.161 & 0.832 & 0.351 & 0.743 & 0.011 & 0.000 & \textbf{0.975} & 0.735 & 0.877 & 0.672 \\
Real EN pages & 0.998 & 0.994 & 0.907 & 0.999 & \textbf{1.000} & 0.999 & 0.746 & 0.862 & 0.087 & 0.836 & 0.858 & 0.604 & 0.431 \\
Real AR page & \textbf{0.993} & 0.979 & 0.601 & 0.986 & \textbf{0.993} & \textbf{0.993} & 0.685 & 0.392 & 0.140 & 0.510 & 0.643 & 0.448 & 0.371 \\
\bottomrule
\end{tabular}
}
\captionof{table}{Mathematical expression recognition: score per test (0--1, higher is better).}
\label{tab:maths-grid}
\end{table*}
\begin{multicols}{2}

Specialist formula systems degraded sharply outside printed formulas: pix2tex
fell from 0.878 on printed formulas to 0.240 on handwritten formulas and 0.020
on Arabic pages; GOT-OCR 2.0 fell from 0.822 to 0.330 and 0.088. Surya returned
no output for 3 of 35 English formulas despite a retry. Counting these failures
as zero gave 0.795, compared with 0.870 over the 32 answered formulas.

Strict checks exposed errors hidden by partial-credit scores (Table~\ref{tab:strict-arabic}). Gemini reproduced
the correct numerical values in 8 of 10 Arabic-notation samples, while digit-form
conversion was also observed in its outputs. Numerical correctness and faithful
digit-form retention were therefore separate diagnostics. On the real Arabic page,
olmOCR 2 and Surya scored 0.993 while replacing an Arabic variable with a Latin
letter; the substitution affected only one token in a page of more than one
hundred tokens.

\begin{center}
\begin{minipage}{\columnwidth}
\centering
\captionof{table}{Strict checks on the ten Arabic-notation samples. ``Symbols''
is the number of samples retaining every symbol; percentages refer to Arabic
letters used as mathematical symbols. The last column reports the strict
Eastern-number retention check, not merely the presence of an Eastern digit.}
\label{tab:strict-arabic}
\scriptsize
\setlength{\tabcolsep}{3pt}
\resizebox{0.96\linewidth}{!}{%
\begin{tabular}{@{}lrrr@{}}
\toprule
System & Symbols & Arabic letters & Eastern numbers \\
\midrule
Azure DI & 8 & 89\% & 8 \\
Document AI & 7 & 100\% & 7 \\
Datalab & 5 & 89\% & 5 \\
Qwen3-VL & 4 & 67\% & 6 \\
Gemini & 3 & 33\% & 3 \\
Surya & 3 & 56\% & 3 \\
Tesseract & 2 & 67\% & 2 \\
dots.ocr & 1 & 67\% & 1 \\
olmOCR 2 & 0 & 78\% & 0 \\
KATIB & 0 & 22\% & 0 \\
PaddleOCR-VL & 0 & 0\% & 1 \\
GOT-OCR 2.0 & 0 & 0\% & 0 \\
pix2tex & 0 & 0\% & 0 \\
\bottomrule
\end{tabular}
}
\end{minipage}
\end{center}

\subsection{Operational Results}

Accuracy was not aligned with speed or hardware cost. Google Document AI was the
fastest text-recognition system at 1.7 seconds per page. The leading self-hosted system,
Surya, required approximately 2~GB of GPU memory but took 19.4 seconds per page
in the single-page test. olmOCR 2 was faster at 12.2 seconds but required about
17~GB of GPU memory.

\begin{center}
\captionof{table}{Operational results for leading text-recognition systems.}
\label{tab:operational}
\scriptsize
\setlength{\tabcolsep}{3pt}
\begin{tabular}{@{}lrrr@{}}
\toprule
System & Sec./page & GPU & Usage fee/1,000 \\
\midrule
Document AI & 1.7 & --- & US\$1.50 \\
Azure DI & 10.4 & --- & ${\sim}$US\$10 \\
olmOCR 2 & 12.2 & 17 GB & US\$0$^{*}$ \\
Surya & 19.4 & 2 GB & US\$0$^{*}$ \\
Datalab & 20.7 & --- & ${\sim}$US\$10 \\
\bottomrule
\end{tabular}
\end{center}

The cloud services required no local GPU but processed documents externally.
Self-hosted systems avoided per-page fees and retained documents locally, but
incurred infrastructure and maintenance costs. The zero-fee entries in
Table~\ref{tab:operational}
exclude GPU operation and any commercial licence. Runtime values reflect
single-page processing and not optimised batch throughput. They compare the
tested pipelines, including their inference backends and API overhead, rather
than architecture-independent speed. Usage fees are not total deployment costs.

\section{Discussion}

The results show that OCR system selection is a multi-objective decision. No
single system was strongest across detection, multilingual text, mathematical
content, speed, and deployment requirements. This variation supports the use of
task-level evaluation rather than a single general ranking, consistent with
recent OCR and document-parsing benchmarks
\cite{liu2023ocrbench,ouyang2025omnidocbench}.

\subsection{Task-Dependent Performance}

The strongest system depended on the target task. Document AI and Surya led
line-level bounding box detection, while Gemini located nearly all text at a
coarser paragraph or cell granularity. The cloud services achieved the lowest
mean CER, although differences between the three were not detected and Surya matched the best cloud median
on a typical page. Mathematical recognition produced a different ranking:
leading self-hosted systems matched the cloud on real English pages, while Azure
DI and Document AI were the strongest group on Arabic notation.

These differences reflect model design. Dedicated OCR services were strong on
speed, print, and page coverage. Document-conversion systems performed better on
complex Arabic material and structured mathematical output. General
vision-language models differed widely: Gemini located text accurately but at
the wrong granularity for line detection, whereas Qwen3-VL was \textit{Poor} on
12 of 14 Arabic and French pages. A system that performs well on one stage of
the pipeline should therefore not be assumed to perform well on the others.

\subsection{Limits of Aggregate Metrics}

Mean CER provided a useful overall comparison, but it did not fully represent
production reliability. PaddleOCR-VL had a median CER of 0.124 but a mean of
0.634, driven by five pages with CER above 1; capping page CER at 1 reduced its
mean to 0.260. The same adjustment moved dots.ocr from eighth to fifth. Median,
capped mean, and failed-page rates therefore distinguish typical performance
from occasional empty, repetitive, or runaway outputs.

Text agreement also did not measure document structure. A system could read all
table values but detach them from their row labels, or return the first rows and
omit the remainder. Layout and table benchmarks similarly distinguish textual
content from structural correctness \cite{pfitzmann2022doclaynet,smock2022pubtables}.
For extraction tasks, row alignment, reading order, and completeness may be more
important than a small difference in CER.

The same limitation appeared in mathematical recognition. On the real Arabic
page, olmOCR 2 and Surya shared a score of 0.993 while replacing an Arabic
variable with a Latin letter. Conversely, equivalent LaTeX structures could
receive different text-based scores despite rendering the same expression.
Normalising matrix environments raised Azure DI from 0.836 to 0.912 and
GOT-OCR 2.0 from 0.862 to 0.940 on the real English pages without changing the
top five systems.
These findings suggest that formula evaluation should combine structural
comparison, strict number checks, and visual or semantic validation, building
on existing image-to-markup and rendering-based evaluation approaches
\cite{deng2017image,wang2025cdm}.

\subsection{Arabic-Specific Reliability}

Arabic heritage material and notation remained the most difficult content.
Datalab led manuscripts and calligraphy, while Surya was the strongest local
model on these document types. The gap between routine print and heritage pages
supports findings from Arabic OCR datasets that handwriting, calligraphy,
degradation, and mixed scripts require separate evaluation
\cite{bhatia2024qalam,ismail2023hicma,heakl2025kitab}.

Arabic mathematics introduced a second distinction: faithful transcription
versus mathematical normalisation. Converting Eastern Arabic digits, variables,
or function names into Latin notation may preserve the mathematical meaning but
does not preserve the document. Gemini reproduced correct number values in 8 of
10 notation samples, while conversion of Eastern Arabic digits to Western digits
was also observed. Its notation score was 0.358; digit-form conversion was assessed separately
because the main notation score mapped Eastern and Western digits to the same
values. This behaviour can be useful for search
or reasoning, but it is unsuitable for archival digitisation without explicit
normalisation rules. Arabic variables and digits should therefore be validated
separately from the rest of the expression.

\subsection{Deployment Implications}

On this benchmark, Surya offered the best balance among self-hosted systems: its
detection was close to Document AI, it had the lowest self-hosted mean CER, and
it required about 2~GB of GPU memory. Its mathematical performance was strong on
full English pages (0.999) but weaker on cropped formulas (0.795, including three
empty outputs) and Arabic notation (0.351). Its model licence is free only for
organisations below the stated revenue and funding thresholds; other use
requires a commercial licence \cite{surya}.

Cloud services were also specialised. Datalab had a clear advantage on Arabic
manuscripts and calligraphy and belonged to the leading group on each
mathematical test except Arabic notation. Azure DI was strongest on Arabic
tables and, with Document AI, on faithful Arabic notation. Document AI combined
the fastest text-processing time, the most consistent detection, no table
truncation, and the lowest listed per-page fee. These findings must be balanced
against external processing and continuing usage fees. Self-hosting retains
documents locally but transfers cost to hardware, maintenance, throughput
planning, and licensing.

\subsection{Limitations}

The document sets were purposefully curated and are not statistically
representative of all institutional collections. The text reference was
produced by a language model rather than transcribed by people, so CER measures
agreement with that reference rather than certified accuracy. Gemini may
normalise digits or punctuation; the stable rankings under stricter
normalisations test sensitivity to preprocessing, not the correctness of the
reference itself. A human-transcribed subset is still needed. The pooled CER
depends on the collection composition, and pages from the same source may be
correlated; the page-level bootstrap did not model that dependence. Per-type
comparisons were exploratory and had no multiple-comparison adjustment.
Detection grades were assigned through one panel-review process, so inter-rater
reliability was not measured and blinding was not documented. Detection findings
apply to the tested prompts and line-box requirement, not to all possible region
granularities. The mathematical subsets
are small, particularly the ten Arabic-notation samples and single real Arabic
page. The generated Arabic-page test used Latin-notation equations and produced
near-ceiling scores for several systems, limiting discrimination on that subset.
Surya returned no output for three English formulas despite a retry.
Runtime was measured one page at a time and may not reflect batched throughput.
Provider prices, regions, versions, and licence terms should be revalidated
before deployment.

\section{Future Work}

Three gaps follow directly from the results. First, a human-transcribed subset,
stratified towards Arabic heritage material, should measure reference error and
whether the ranking holds. Second, more real Arabic mathematics pages are needed
because the single real page already exposed variable substitution. Third,
detection should be scored at both line and paragraph granularity because
Gemini's result depends on which granularity is required. Independent raters,
document-source-aware resampling, and language- and type-balanced summaries
should accompany these extensions. The expanded set
should also include more handwriting, calligraphy, tables, and mixed-script
pages \cite{ismail2023hicma,heakl2025kitab}.

The evaluation should extend beyond character agreement. Future metrics should
measure reading order, table structure, bounding-box overlap, formula rendering,
and semantic equivalence. Strict checks for Arabic digits, variables, and
notation should remain separate because small substitutions can change meaning.
Layout-aware and rendering-based measures provide useful foundations for this
extension \cite{smock2022pubtables,ouyang2025omnidocbench,wang2025cdm}.

Further experiments should evaluate batch throughput, energy use, long-document
processing, and repeated runs across model versions. Hybrid pipelines that route
pages by language, layout, or content type should also be compared with single
systems. Such work would determine whether task-specific combinations improve
accuracy while satisfying cost, privacy, and hardware constraints.

\section{Conclusion}

Document AI and Surya produced \textit{Good} line boxes on 98\% and 96\% of
pages, while Gemini located text accurately at a coarser granularity. The three
cloud services showed no detected difference in the exploratory text-recognition
comparisons; Surya had the
lowest mean CER among self-hosted systems, with the largest remaining gaps on
heritage material and tables. Available paired comparisons did not establish a
mathematical leader on any test, and high
page scores concealed replaced Arabic variables and rewritten digits. OCR
selection for Arabic collections should therefore combine task-level accuracy
and uncertainty with strict checks of Arabic symbols and institutional
deployment constraints.

\section*{Declarations}

\textbf{Conflict of Interest.} The authors declare no conflicts of interest.

\textbf{Acknowledgment.} The authors acknowledge the role of the Naseej
Innovation Lab research team in supporting the development and evaluation of
this benchmark.

\bibliographystyle{plainnat}
\bibliography{references}

\end{multicols}

\appendix

\section{Systems and Settings}

Self-hosted systems ran on one NVIDIA RTX A6000 (48~GB) with Python 3.11,
PyTorch 2.7, and Transformers 5.18. Cloud services were called through their
APIs at temperature 0 where that setting was available. Table~\ref{tab:settings}
records the identifiers and principal inference settings used in the benchmark.
The bibliography links to research papers, technical reports, and primary
project or provider documentation. Model cards and service documentation identify
the evaluated systems; they are not independent evidence of comparative
accuracy. The listed identifiers do not substitute for archived checkpoint
revisions or immutable service versions.

\begin{center}
\begin{minipage}{\textwidth}
\centering
\captionof{table}{Evaluated systems and principal settings.}
\label{tab:settings}
\scriptsize
\begin{tabular}{@{}p{0.18\textwidth}p{0.30\textwidth}p{0.44\textwidth}@{}}
\toprule
System & Model or service & Settings \\
\midrule
Gemini 3.1 Flash-Lite & google/gemini-3.1-flash-lite via OpenRouter & Temperature 0; image long edge 1,600 px; line boxes as $[y_{\min},x_{\min},y_{\max},x_{\max}]$ on 0--1000; LaTeX-oriented mathematical prompt \\
Qwen3-VL & Qwen/Qwen3-VL-8B-Instruct & bfloat16; maximum $1280\times28\times28$ pixels; line boxes as $[x_1,y_1,x_2,y_2]$ on 0--1000 \\
olmOCR 2 & allenai/olmOCR-2-7B-1025 & bfloat16; greedy decoding; repetition penalty 1.15; up to 1,536 new tokens \\
dots.ocr & rednote-hilab/dots.ocr & Layout-and-text JSON prompt \\
PaddleOCR-VL & PaddlePaddle/PaddleOCR-VL (0.9B) & fp32 \\
KATIB & oddadmix/Katib-Qwen3.5-0.8B-0.1 & Verbatim transcription prompt; diacritics retained \\
Surya & Surya OCR 2 (GGUF; llama.cpp server) & Text and LaTeX modes; Surya detection model for line boxes \\
GOT-OCR 2.0 & stepfun-ai/GOT-OCR-2.0-hf & Formatted LaTeX output for mathematics \\
pix2tex & LaTeX-OCR & Default checkpoint \\
Tesseract & Tesseract 4.1.1 & ara, fra, and eng language packs \\
Datalab & Datalab API & Accurate mode \\
Azure DI & prebuilt-layout, API 2024-11-30 & Formula add-on for mathematics \\
Document AI & Enterprise Document OCR & US region \\
\bottomrule
\end{tabular}
\end{minipage}
\end{center}

For detection, prompts requested one box per text line and the coordinate order
shown above. Text prompts requested verbatim transcription without commentary.
Mathematical prompts requested LaTeX that preserved the written symbols, digit
forms, variables, and reading order. Outputs were retained without manual repair.

\section{Normalisation and Scoring Rules}
\label{app:normalisation}

This appendix specifies the rules used to prepare texts and compute the scores
reported in the paper.

\subsection{Text Recognition}

\subsubsection{Markup Removal}

The same steps were applied to each system output and its reference:
\begin{enumerate}
    \item Markdown images were removed, and Markdown links were replaced by
    their visible text.
    \item HTML line breaks and the ends of table cells, rows, paragraphs, and
    headings became spaces or line breaks. Other HTML tags were removed, and
    HTML entities were decoded; for example, \texttt{\&amp;} became \texttt{\&}.
    \item Markdown table separator rows and table pipes (\texttt{|}) were removed.
    \item Markdown heading markers (\texttt{\#}) were removed.
    \item Bold markers (\texttt{**} and \texttt{\_\_}) were removed, retaining
    their text. Single asterisks, as in \texttt{3*4}, were kept.
\end{enumerate}

\subsubsection{Text Normalisation}

Both texts were converted to Unicode Normalisation Form C (NFC). Every run of
whitespace, including spaces, tabs, and line breaks, became a single space;
leading and trailing spaces were removed. Nothing else was changed in the base
comparison: case, punctuation, Arabic diacritics, and Eastern Arabic digits
were retained as written.

\subsubsection{Character Error Rate and Aggregate Measures}

For each page, CER was the character-level Levenshtein distance between output
and reference divided by the number of reference characters. Insertions,
deletions, and substitutions each had unit cost. CER was zero for identical
texts and could exceed 1 when an output was much longer than its reference.
Mean and median CER were calculated across scored pages. Capped mean CER limited
each page value to 1 before averaging.

Confidence intervals used a 95\% percentile bootstrap with 2,000 page resamples
and random seed 42. Paired comparisons bootstrapped per-page CER differences
on pages scored by both systems. A difference was reported as clear when its
95\% interval excluded zero.

\subsubsection{Supplementary Measures and Robustness Checks}

Word error rate used the same edit-distance procedure on space-separated words.
For CER without diacritics, Arabic diacritics in U+064B--U+0652,
U+0653--U+0655, and U+0670, together with tatweel (U+0640), were removed from
both texts. Diacritic error rate was CER on the diacritic sequence alone, for
pages whose reference contained at least five diacritics.

Robustness checks repeated text scoring with letters only, diacritics removed,
Eastern and Western digits treated as equivalent, and punctuation removed.
Kendall's $\tau$ against the base ranking was 1.00, 0.96, 0.96, and 1.00,
respectively.

\subsubsection{Output Problems}

Each output was checked after NFC and whitespace normalisation according to
Table~\ref{tab:failure-definitions}. A failed page had at least one problem;
categories could overlap. Requests that failed for technical reasons on the
evaluation side were not scored.

\begin{center}
\begin{minipage}{\textwidth}
\centering
\captionof{table}{Definitions of text-recognition output problems.}
\label{tab:failure-definitions}
\small
\begin{tabular}{@{}p{0.20\linewidth}p{0.74\linewidth}@{}}
\toprule
Problem & Definition \\
\midrule
Empty & No text returned. \\
Caption & The output described the image instead of transcribing it, using
phrases such as ``this image shows'', ``appears to be'', ``I'm sorry'',
or ``as an AI''. \\
Repetition loop & More than half of the six-word sequences in the output
were repeats; tested on outputs containing at least 12 words. \\
Very short & Fewer than 30 characters. \\
Language switch & On an Arabic or mixed-language page, more than 20 letters
were returned and Latin letters outnumbered Arabic letters. \\
No Arabic output & On an Arabic or mixed-language page, no Arabic letters
were returned. \\
\bottomrule
\end{tabular}
\end{minipage}
\end{center}

dots.ocr had one unscored English table page; its aggregate statistics used 314
scored pages.

\subsection{Mathematical Expression Recognition}

\subsubsection{Equation Extraction}

Equations were extracted from \verb|$...$|, \verb|$$...$$|,
\verb|\(...\)|, \verb|\[...\]|, or an \texttt{equation}, \texttt{align},
\texttt{gather}, or \texttt{displaymath} environment. HTML \texttt{<math>}
content was retained as LaTeX; other HTML was removed. If these markers were
absent, lines containing a LaTeX command, \verb|^|, \verb|_|, or a mathematical
symbol were used, provided that no more than 20\% of their characters were
Arabic.

\subsubsection{LaTeX Normalisation}

The following operations were applied in order:
\begin{enumerate}
    \item Dollar signs were removed.
    \item Functions inside font commands were standardised;
    for example, \verb|\mathrm{s i n}| became \verb|\sin|. This covered
    \texttt{sin}, \texttt{cos}, \texttt{tan}, \texttt{log}, \texttt{ln},
    \texttt{exp}, \texttt{lim}, \texttt{max}, \texttt{min}, \texttt{det},
    \texttt{gcd}, and similar operators. \verb|\operatorname{f}| became \verb|\f|.
    \item Size and delimiter commands were dropped: \verb|\left|,
    \verb|\right|, \verb|\big|, \verb|\Big|, \verb|\bigg|, \verb|\Bigg|,
    \verb|\displaystyle|, \verb|\textstyle|, \verb|\scriptstyle|,
    \verb|\limits|, \verb|\nolimits|, \verb|\quad|, and \verb|\qquad|.
    \item Spacing commands were dropped: \verb|\,|, \verb|\:|, \verb|\;|,
    \verb|\!|, \verb|\>|, a backslash followed by a space, and \verb|~|.
    \item Font commands were removed while retaining their content:
    \verb|\mathrm|, \verb|\textrm|, \verb|\text|, \verb|\mathit|,
    \verb|\rm|, and \verb|\mbox|.
    \item Equivalent commands were aliased as in Table~\ref{tab:latex-aliases}.
\end{enumerate}

\begin{center}
\begin{minipage}{0.70\textwidth}
\centering
\captionof{table}{Equivalent LaTeX command aliases.}
\label{tab:latex-aliases}
\small
\begin{tabular}{@{}ll@{}}
\toprule
Written as & Treated as \\
\midrule
\verb|\dfrac|, \verb|\tfrac| & \verb|\frac| \\
\verb|\dbinom|, \verb|\tbinom| & \verb|\binom| \\
\verb|\le|, \verb|\ge|, \verb|\ne| & \verb|\leq|, \verb|\geq|, \verb|\neq| \\
\verb|\to|, \verb|\gets| & \verb|\rightarrow|, \verb|\leftarrow| \\
\verb|\lvert|, \verb|\rvert|, \verb|\vert| & \verb!|! \\
\bottomrule
\end{tabular}
\end{minipage}
\end{center}

\subsubsection{Tokenisation and Token Score}

Whitespace was ignored. Each LaTeX command or escaped character, brace,
\verb|^|, \verb|_|, \verb|&|, single letter, single digit, or other symbol
formed one token. Braces around a single token and empty brace pairs were
removed repeatedly until no further changes occurred. For example,
\verb|\frac{2}{x}| and \verb|\dfrac 2x| became the same token sequence.

For output token sequence $T_o$ and reference sequence $T_r$, the score was
\begin{equation}
    s_{\mathrm{token}}=
    \max\left(0,1-\frac{d_{\mathrm{Lev}}(T_o,T_r)}{|T_r|}\right).
\end{equation}
Here $d_{\mathrm{Lev}}$ is token-level Levenshtein distance. Identical sequences
scored 1, while partly correct equations received partial credit.

\subsubsection{Main Score for Each Test}

\textbf{English formulas.} Extracted equations were joined and compared with
the reference LaTeX using the token score. If no equation markers were present,
the whole output was used after HTML removal.

\textbf{Arabic pages.} The score averaged two components: $1-\mathrm{CER}$ on
text outside equations, after NFC conversion, whitespace collapse, and mapping
Eastern Arabic digits to 0--9; and the token score of the equations.

\textbf{Arabic notation.} The score was $1-\mathrm{CER}$ on the whole equation
after HTML removal, NFC conversion, mapping digits in U+0660--U+0669 and
U+06F0--U+06F9 to 0--9, and removal of all spaces. Thus the main score compared
digit values, while preservation of the written digit form was checked
separately.

\textbf{Real pages.} The whole page was compared as one token sequence after
removing HTML, all equation markers, layout environments (\texttt{aligned},
\texttt{align}, \texttt{gathered}, \texttt{split}, \texttt{array},
\texttt{equation}, and \texttt{displaymath}), \verb|&|, \verb|\\| line breaks,
and Markdown headings and bold markers. Correctly reproduced pages scored 1
when the resulting normalised token sequence matched the reference exactly;
incorrect numbers lost credit. Equivalent delimiters and the listed layout
commands were removed, but other equivalent LaTeX forms could still produce
different token sequences. Matrix-environment normalisation was assessed
separately as a sensitivity check, not assumed to eliminate all style effects.

\subsubsection{Additional Checks on Arabic Notation}

\begin{itemize}
    \item \textbf{Numbers right:} all digits in order matched the reference after
    Eastern Arabic digits were mapped to 0--9.
    \item \textbf{Kept Arabic digits:} the output contained Eastern Arabic digits
    when the reference did.
    \item \textbf{Converted to Latin:} the output contained a LaTeX command,
    \verb|^|, \verb|_|, or a Latin variable adjacent to a number or operator,
    as in \texttt{3x} or \texttt{x = 1}.
\end{itemize}
These diagnostics complemented the strict symbol-retention checks reported in
Table~\ref{tab:strict-arabic}. The presence-based ``kept Arabic digits'' check
was weaker than strict retention of all written Eastern numbers: an output
could mix Eastern and Western digits. These indicators were not complementary
counts and should not be subtracted from the number of samples to infer a
conversion rate. Empty outputs scored zero; requests that failed
for technical reasons on the evaluation side were not scored. Confidence
intervals used the bootstrap procedure in Appendix~B.1.

\subsection{Failure Types}

Table~\ref{tab:failure-types} reports the share of scored text pages with each
output problem defined in Appendix~B.1. The final column gives the share with
at least one problem. Categories overlap and must not be summed.

\begin{center}
\begin{minipage}{\textwidth}
\centering
\captionof{table}{Text-recognition failure types on 315 scored pages; dots.ocr
uses 314 scored pages. A page can have more than one problem.}
\label{tab:failure-types}
\small
\setlength{\tabcolsep}{4pt}
\begin{tabular}{@{}lrrrrrrr@{}}
\toprule
System & Empty & Repetition & Language & No Arabic & Very short & Caption & Any \\
 & & loop & switch & output & & & (failed) \\
\midrule
PaddleOCR-VL & 1.3\% & 1.9\% & 6.3\% & 0.3\% & 0.6\% & 0\% & 9.5\% \\
dots.ocr & 1.6\% & 4.1\% & 1.0\% & 0\% & 0.3\% & 0\% & 7.0\% \\
KATIB & 0\% & 5.1\% & 0.3\% & 0\% & 0.3\% & 0\% & 5.4\% \\
Tesseract & 3.5\% & 0\% & 0\% & 0\% & 1.6\% & 0\% & 5.1\% \\
Qwen3-VL & 0.3\% & 2.9\% & 0\% & 0\% & 0\% & 0\% & 3.2\% \\
Azure DI & 0\% & 0\% & 2.2\% & 0.3\% & 0\% & 0\% & 2.2\% \\
Surya & 1.0\% & 0.6\% & 0.3\% & 0.3\% & 0.3\% & 0\% & 2.2\% \\
Document AI & 0\% & 0\% & 0.6\% & 0.6\% & 0\% & 0\% & 0.6\% \\
Datalab & 0\% & 0\% & 0.6\% & 0.6\% & 0\% & 0\% & 0.6\% \\
olmOCR 2 & 0\% & 0.3\% & 0.3\% & 0\% & 0\% & 0\% & 0.6\% \\
\bottomrule
\end{tabular}
\end{minipage}
\end{center}

\section{Example Documents Used in the Experiments}
\label{app:document-examples}

\begin{figure}[!htbp]

\centering
\begin{minipage}[t]{0.48\textwidth}
    \centering
    \includegraphics[width=\linewidth]{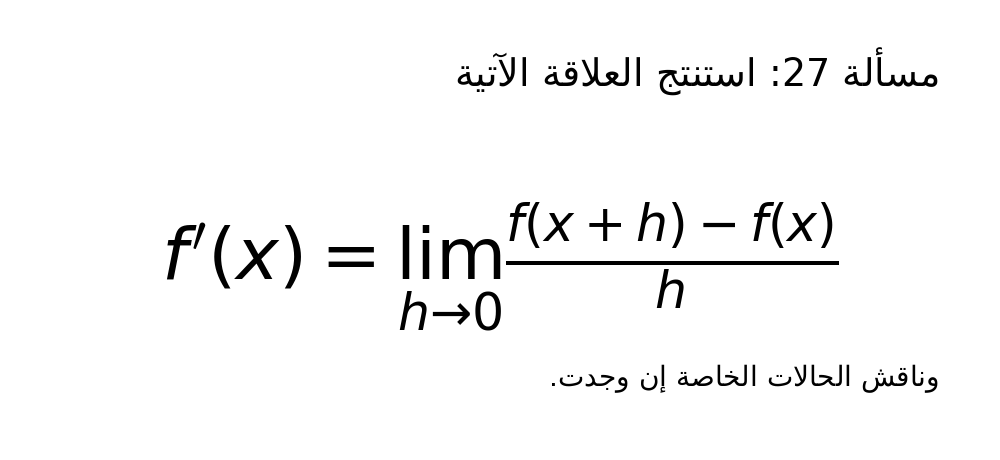}
    \caption*{(a) Arabic mathematical sample (MATH-AR-027).}
\end{minipage}\hfill
\begin{minipage}[t]{0.48\textwidth}
    \centering
    \includegraphics[width=\linewidth]{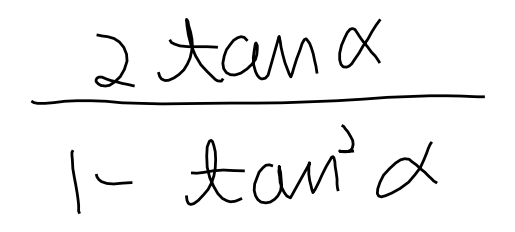}
    \caption*{(b) English mathematical sample (MATH-EN-031).}
\end{minipage}

\vspace{1em}

\begin{minipage}[t]{0.48\textwidth}
    \centering
    \includegraphics[width=\linewidth]{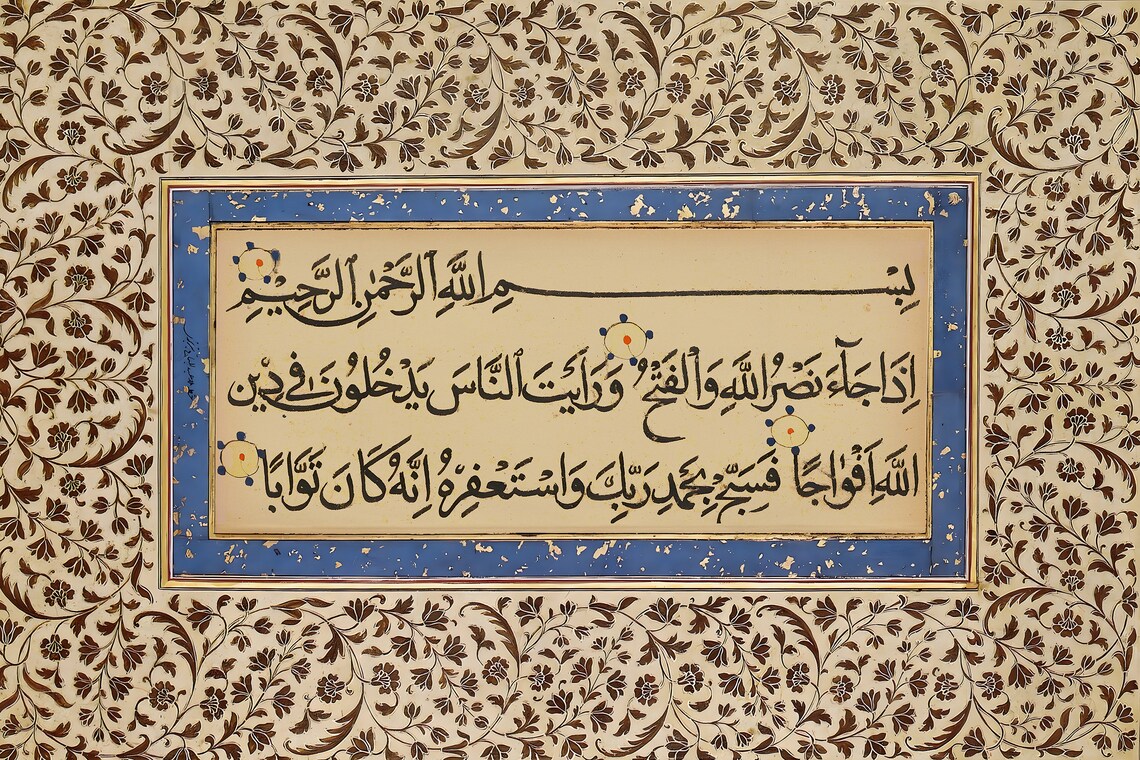}
    \caption*{(c) Arabic calligraphy sample (AR-CALLI-006).}
\end{minipage}\hfill
\begin{minipage}[t]{0.48\textwidth}
    \centering
    \includegraphics[width=\linewidth]{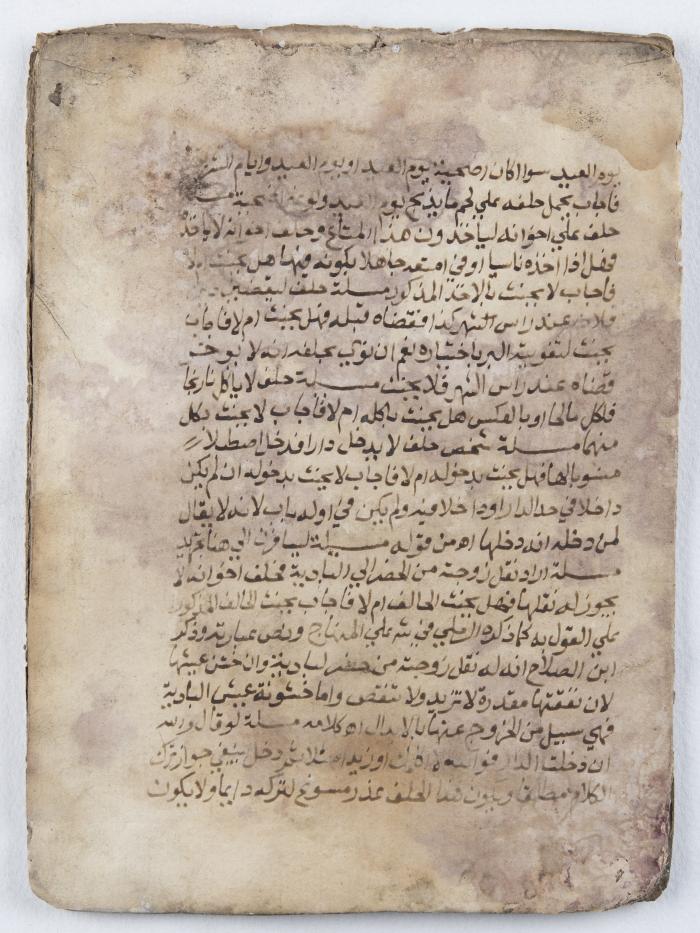}
    \caption*{(d) Arabic manuscript sample (AR-MANU-004).}
\end{minipage}
\caption{Examples of documents included in the mathematical-recognition and
Arabic calligraphy and manuscript experiments.}
\label{fig:document-examples}
\end{figure}

Figure~\ref{fig:document-examples} shows representative Arabic and English
mathematical samples and Arabic calligraphy and manuscript documents used in
the experiments.

\end{document}